%% file: paper.tex
\documentclass[runningheads]{llncs}

\usepackage[utf8]{inputenc}
\usepackage[hidelinks]{hyperref}
\usepackage{graphicx}
\usepackage{blindtext}
\usepackage{url}
\usepackage{subcaption}
\usepackage{verbatimbox}
\usepackage{amssymb}
\usepackage{amsmath}
\usepackage{cleveref}
\usepackage{xcolor}
\usepackage{wrapfig}

\crefformat{footnote}{#2\footnotemark[#1]#3}

\begin{document}

\mainmatter

\title{Inflection-Tolerant Ontology-Based Named Entity Recognition for Real-Time Applications}

\titlerunning{Inflection-Tolerant Ontology-Based NER for Real-Time Applications}

\author{
  Christian Jilek\inst{1,2} \and
  Markus Schröder\inst{1,2} \and
  Rudolf Novik\inst{2} \and
  Sven Schwarz\inst{1} \and\\
  Heiko Maus\inst{1} \and
  Andreas Dengel\inst{1,2}
}

\authorrunning{C. Jilek et al.}

\institute{
  Smart Data \& Knowledge Services Department, DFKI GmbH,\\Kaiserslautern, Germany\\
  \email{\{christian.jilek,markus.schroeder,sven.schwarz,\\heiko.maus,andreas.dengel\}@dfki.de}\\ \and
  Computer Science Department, TU Kaiserslautern, Kaiserslautern, Germany\\
  \email{r\_novik15@cs.uni-kl.de}
}

\maketitle

\begin{abstract}
\input{content/frontback_abstract}
\end{abstract}

\input{content/main_intro}
\input{content/main_relatedwork}
\input{content/main_approach}
\input{content/main_evaluation}
\input{content/main_conclusion}

\input{content/frontback_acks}

\bibliographystyle{splncs04}
\bibliography{content/refs}

\end{document}

%% file: content/frontback_abstract.tex
A growing number of applications users daily interact with have to operate in (near) real-time:
chatbots, digital companions, knowledge work support systems -- just to name a few.
To perform the services desired by the user, these systems have to analyze user activity logs or explicit user input extremely fast.
In particular, text content (e.g. in form of text snippets) needs to be processed in an information extraction task.
Regarding the aforementioned temporal requirements, this has to be accomplished in just a few milliseconds, which limits the number of methods that can be applied.
Practically, only very fast methods remain, which on the other hand deliver worse results than slower but more sophisticated Natural Language Processing (NLP) pipelines.

In this paper, we investigate and propose methods for real-time capable Named Entity Recognition (NER).
As a first improvement step we address are word variations induced by inflection, for example present in the German language.
Our approach is ontology-based and makes use of several language information sources like Wiktionary.
We evaluated it using the German Wikipedia (about 9.4B characters), for which the whole NER process took considerably less than an hour.
Since precision and recall are higher than with comparably fast methods, we conclude that the quality gap between high speed methods and sophisticated NLP pipelines can be narrowed a bit more without losing too much runtime performance.

\keywords{
  Ontology-based information extraction \and
  Named entity recognition \and
  Inflectional languages \and
  Real-time systems
}

%% file: content/main_intro.tex
\section{Introduction}
\label{sec:introduction}
\noindent
The number of application areas, in which users are supported by systems that operate in (near) real-time, grows: chatbots, digital companions, knowledge work support systems -- just to name a few.
Our particular scenario involves a system based on Semantic Desktop  \cite{SauermannBernardiDengel2005} technology, that semi-automatically re-organizes itself based on user context in order to better support knowledge work and information management activities \cite{JilekSchroederSchwarz+2018}.
We envision an intelligent, proactive assistance parallel to the actual work.
Such systems need mechanisms to analyze observed user activities (entering text, browsing a website, reading/writing files, \ldots) in order to decide on the right support measures and perform them accordingly.
The demand for very short reaction times limits the number of methods that can be applied.

In this paper, we focus on Information Extraction (IE) methods, more precisely Named Entity Recognition (NER), that are ontology-based (our system operates on knowledge graphs in the background) and meet the demand for providing meaningful results within only a few milliseconds on users' typical computing devices.
By \textit{only a few} we actually mean a small two-digit number of milliseconds.
According to Miller (1968) and Card et al. (1991), as cited in \cite{Nielsen1993}, 100 ms is ``about the limit for having the user feel that the system is reacting instantaneously'' and 1000 ms is ``about the limit for the user's flow of thought to stay uninterrupted''.
Our goal is to stay below the first value.
In cases, in which this is not possible (e.g. too much data to be processed at once), 1000 ms should be the upper bound of processing time to be tolerated.
Since we also need some time for selecting and performing the support measures, the IE task has to completed within only a fraction of this time span.
Such strict temporal requirements usually rule out very sophisticated Natural Language Processing (NLP) pipelines (higher quality solutions but slow), leaving only rather simple (lower quality) but very fast methods often based on pre-defined rules or gazetteers.
A gazetteer is conceptually just a list of terms (typically static), that the input text is later scanned for, e.g. the names of persons, organizations or locations.
Since our scenario also involves highly inflectional languages like German%
\footnote{other inflectional languages: Spanish, Latin, Hebrew, Hindi, Slavic languages, \ldots}, we additionally have to take slight variations of such terms into account.
To vividly illustrate the problem of inflections in NER, we fed the first paragraph of the German Wikipedia article of \textit{Propositional calculus} (German: \textit{Aussagenlogik}) to \textit{DBpedia Spotlight}%
\footnote{\url{https://www.dbpedia-spotlight.org/demo/}}
\cite{MendesJakobGarciaSilva+2011}, a well-known and often used recognizer for Wikipedia/DBpedia%
\footnote{\url{https://wiki.dbpedia.org/}} entities in given text snippets.
The results are depicted in Figure \ref{fig:spotlight} (middle section):
\begin{figure}[t]
  \centering
  \includegraphics[width=1\columnwidth]{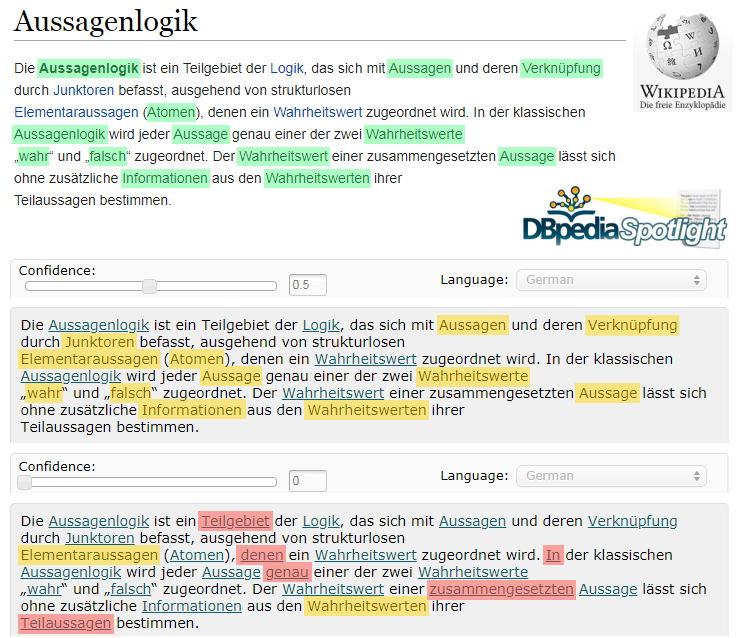}
  \caption{First paragraph of the German Wikipedia article of \textit{Aussagenlogik} (top) fed to DBpedia Spotlight using confidence values of 0.5 (middle) and 0.0 (bottom) (highlighting we applied: green: existing Wikipedia articles not linked in the original document, yellow: false negatives, red: false positives)}
  \label{fig:spotlight}
\end{figure}
Twelve entities (in just three sentences; we highlighted them in yellow) are not found, ten of them due to lexical variations induced by inflection.
E.g. \textit{Wahrheitswert} (\textit{truth value}) is found, whereas its inflected forms ending with \textit{-e} and \textit{-en} are not.
If we lower the confidence to 0.0, there are still some entities missing and false positives come up.

In summary, our goal is to find or implement methods that are fast enough to meet the aforementioned temporal constraints while at the same time achieving better results than standard high speed methods.
Recognizing entities despite the just mentioned lexical variations induced by inflection would be a first improvement step.
Note that disambiguation as well as recognizing Named Entities (NE) yet unknown to the system (i.e. not available as instances in the knowledge graph) are out of this paper's scope.
Since there is a lot of explicitated contextual information available in our system \cite{JilekSchroederSchwarz+2018}, we intend to address disambiguation in our scenario in a future paper.

The rest of this paper is structured as follows:
Section \ref{sec:relatedwork} provides an overview of related work in this area.
Our approach is described in Section \ref{sec:approach} and its evaluation is presented in \ref{sec:evaluation}.
In Section \ref{sec:conclusion}, we conclude this paper and give a an outlook on possible future work.

%% file: content/main_relatedwork.tex
\section{Related Work}
\label{sec:relatedwork}
\noindent
We were looking for approaches (more or less) explicitly addressing inflection tolerance or real-time capability, preferably both at the same time:

Concerning real-time capability, Dlugolinsky et al. \cite{DlugolinskyNguyenLaclavik+2013} present an overview of different gazetteer-based approaches, especially referring to various versions included in the GATE (General Architecture for Text Engineering) framework \cite{CunninghamTablanRoberts+2013}.
They distinguish between character- and token-based variants and state that the latter usually have ``longer running time and low processing performance''.
They thus focus on character-based gazetteers and present several versions \cite{DlugolinskyNguyenLaclavik+2013,NguyenDlugolinskyLaclavik+2014}.
Since some of their implementations are available online, we also included them in our evaluation (see Section \ref{sec:evaluation}).

Savary \& Piskorski \cite{SavaryPiskorski2010} investigated solutions for Polish, also a highly inflectional language.
As one subcomponent of their IE platform \textit{SProUT} they filled a gazetteer by ``explicitly listing all inflected forms of each entry''.

Day \& Prukayastha \cite{DeyPrukayastha2013} gave an overview of different NER methods especially targeting Indian languages.
Their paper presented gazetteer-based and machine learning approaches as well as hybrid solutions.

Al-Jumaily et al. \cite{AlJumailyMartinezMartinezFernandez+2012} present an NER system for Arabic text mining.
They use a token-based approach involving stemming as well as pre- and postfix verification tailored to the Arabic language.
Although they aim for real-time applications, they do not give any details about their system's runtime performance.

Al-Rfou \& Skiena \cite{AlrfouSkiena2012} propose \textit{SpeedRead}, an NER pipeline which they tested to run ten times faster than the \textit{Stanford CoreNLP} pipeline%
\footnote{\url{https://stanfordnlp.github.io/CoreNLP/}}.
Unfortunately for us, they only reported runtime performance in terms of tokens per second.
In their final results, they say SpeedRead achieves about 153 tokens/sec.
Using the word length statistics published by Norvig \cite{Norvig2013} and assuming an average token length of about five characters, we would end up having 765 char./sec, which is still much too slow for our scenario as we will later see.
Even if we assume an average token length of twelve, although more than 90\% of all English words are shorter \cite{Norvig2013}, we would still be too slow having 1836 char./sec.

In summary, we found several approaches dealing with either real-time capability or inflection tolerance.
One paper even mentioned both, but did not report any concrete speed measures.
Nevertheless, doing NER extremely fast is apparently rarely discussed in literature, yet.
This may be because usual NER methods operate in only a few seconds, which may be sufficient for many use cases, unfortunately not ours.

We will refer to some of the ideas discussed in this section when presenting our approach in the following.

%% file: content/main_approach.tex
\section{Approach}
\label{sec:approach}
\noindent
We focus on the very fast recognition of NEs given as instance labels of ontologies.
Moreover, these labels should still be recognized even if they slightly lexically vary as induced by inflection.
To achieve this, we exploit knowledge graphs connected or available to our system such as an individual user's Personal Information Model (PIMO) \cite{SauermannVanElstDengel2007} or DBpedia to get more details about the entities, e.g. their specific type.
Based on this type, we can then accept different lexical variations per instance according to language information coming from Wiktionary%
\footnote{\url{https://www.wiktionary.org/}}, for example.
For instance, we should not allow too many variations of person names, whereas we can be more tolerant when dealing with topic, project, organization or location names, especially if they contain adjectives like the \textit{Technical} University of Kaiserslautern or \textit{German} Research Center for \textit{Artificial} Intelligence.
As an example, Figure \ref{fig:fstht} shows all 18 inflected forms of \textit{künstlich} (\textit{artificial}) in German (word \texttt{w4} in the figure).

\begin{figure}
  \centering
  \includegraphics[width=1\columnwidth]{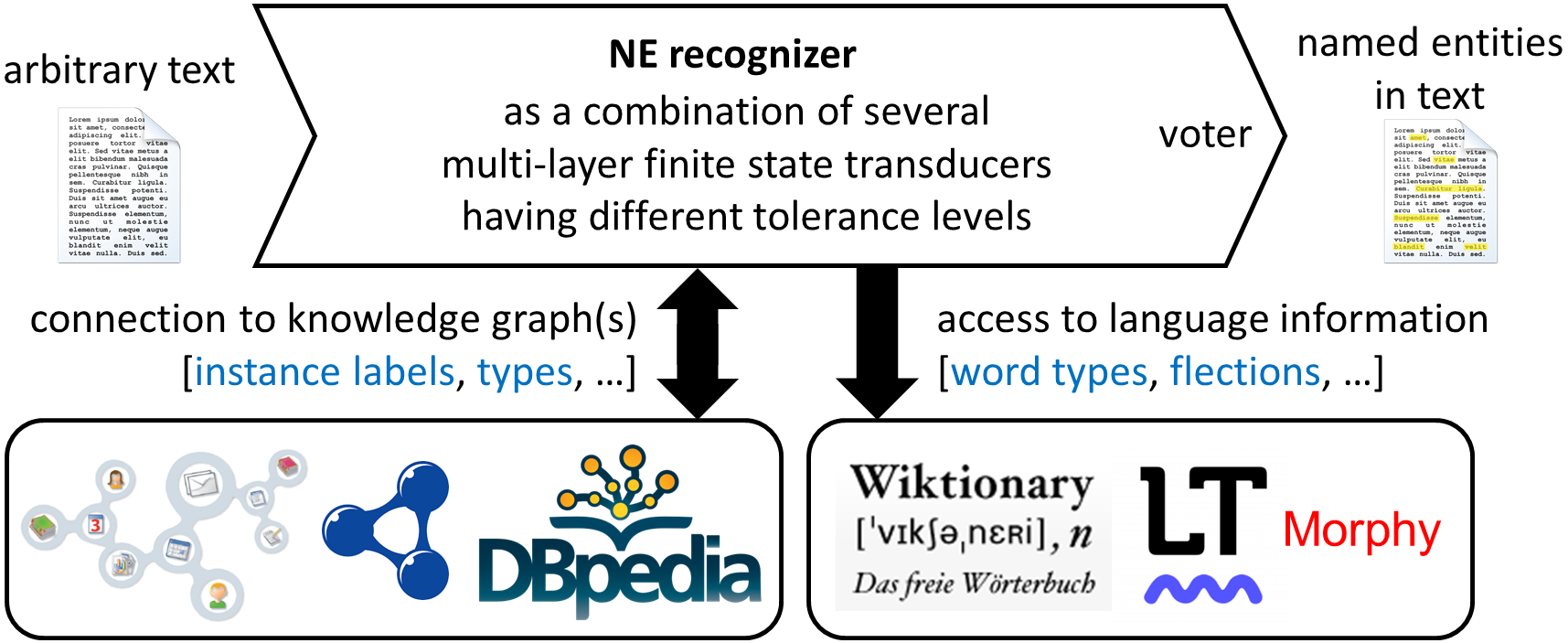}
  \caption{Architecture of our system}
  \label{fig:architecture}
\end{figure}

As depicted in Figure \ref{fig:architecture}, we have a hierarchical NE recognizer as the core of our system.
It operates on several sub-recognizers, mostly Multi-Layer Finite-State Transducers (MLFST) as described later, each of them having a different focus (configuration).
The core recognizer collects their results and decides (votes) which ones to accept.
To acquire the entity labels as well as background information, it is connected to knowledge graphs and language information sources as described before.
Its individual aspects are discussed in more detail in the following.
%

\noindent\textbf{FST-based NER.}
To meet the aforementioned strict runtime requirements, we basically follow a gazetteer-based approach.
The additionally required inflection tolerance is not well compatible with the usually static character of a gazetteer.
We thus need enhancements as described in the following.

Our core method is based on the well-known string matching algorithm by Aho \& Corasick \cite{AhoCorasick1975}.
It operates on \textit{tries}, i.e. trees whose nodes represent characters, which are traversed synchronously to the processing of each character of the input text.
Whenever the traversal ends in an accepting state, there is a string match.
Since, in our case, these strings are the labels of NEs, we additionally demand that their ID or URI is returned, which makes the system a Finite-State Transducer (FST).
The algorithm basically has linear runtime complexity as discussed later.
Our scenario involves a highly dynamic, evolving knowledge graph, in which instances (and especially their labels) can be added, deleted or updated potentially several times per minute.
We thus omitted further optimizations like suffix compression in favor of a fast and easy to update FST structure.
\\

\noindent\textbf{Multi-Layer FST.}
For runtime performance reasons we decided against sophisticated NLP pipelines (test results and more details in Section \ref{sec:evaluation}) and therefore follow the approach of explicitly listing all inflected forms of an entity label as proposed in \cite{SavaryPiskorski2010}.
Without further ado, this would easily lead to memory performance problems due to a considerable increase of the FST, especially for multi-word terms:
The more words such a multi-word term consists of, the more potential combinations exist.
Although inflection tolerance is discussed more thoroughly in the paragraph after next, let us just consider a short example here:
If we allow each combination of inflected forms of the term \textit{Deutsches Forschungszentrum für Künstliche Intelligenz} (\textit{German Research Center for Artificial Intelligence}, shortly referred to as DFKI), although lots of them are grammatically not correct (as also discussed later), we would end up with 576 variations ($=6\cdot3\cdot1\cdot16\cdot2$; see upper part of Figure \ref{fig:fstht}).
Inspired by Abney, who proposed the idea of \textit{finite-state cascades} \cite{Abney1996}, we therefore chose to introduce an additional layer to separate character from word processing, making our system a multi-layer FST as illustrated in Figure \ref{fig:fstht}:
\begin{figure}[t]
  \centering
  \includegraphics[width=0.96\columnwidth]{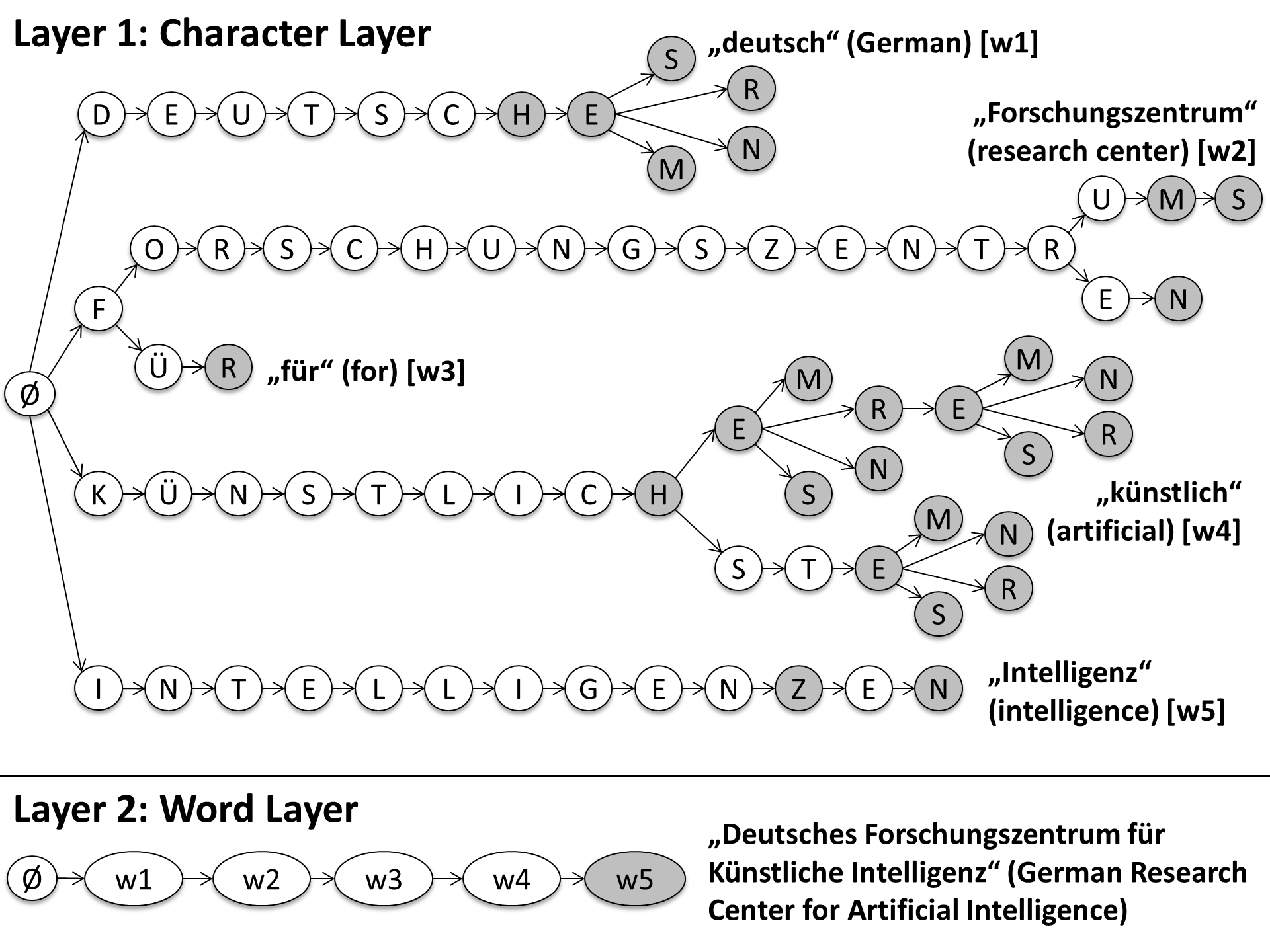}
  \caption{Multi-layer finite transducer consisting of a character and a word layer, and fed with the term \textit{Deutsches Forschungszentrum für Künstliche Intelligenz} ($\varnothing$: start nodes; $w_i$: word IDs; gray nodes: accepting states).}
  \label{fig:fstht}
\end{figure}
Once a word is identified in the first layer (i.e. the FST is in an accepting state; gray node), its ID is passed to the second layer, which checks whether this word may be accepted at this position, either as a single-word or part of a multi-word term.
If a term match is detected, its ID/URI is returned.
As a consequence, each word and its inflected forms, no matter how often or at which positions (in multi-word terms) they appear, only exist once in the FST, thus preventing it from growing too fast in size.

To avoid backtracking in the word layer, the system processes several options in parallel as shown in Figure \ref{fig:rake}:
\begin{figure}[t]
  \centering
  \includegraphics[width=1\columnwidth]{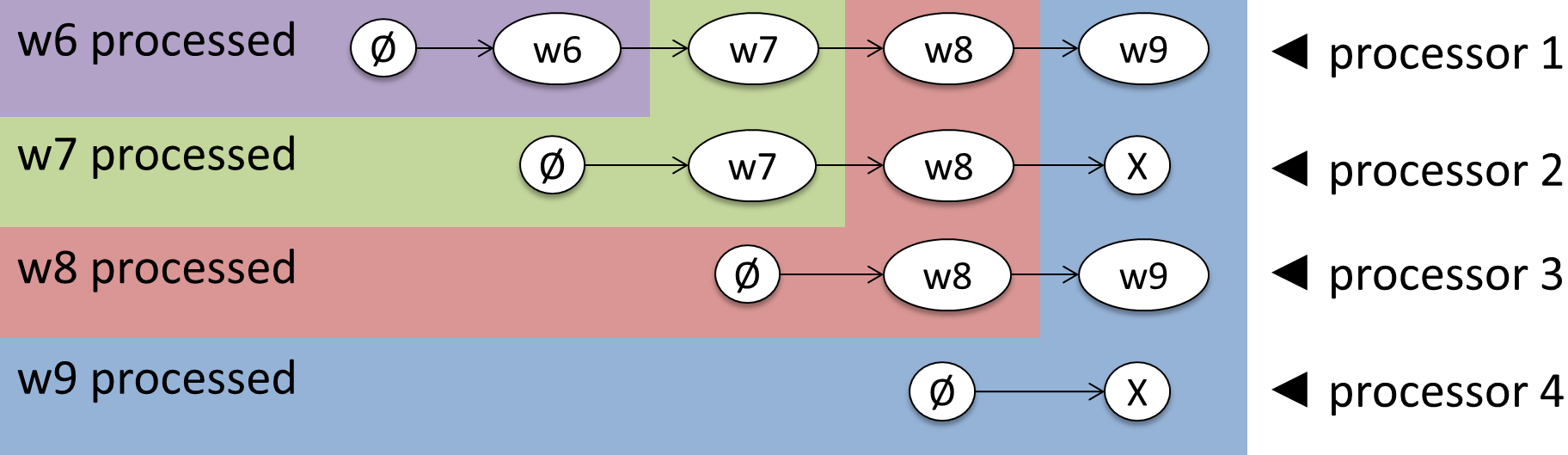}
  \caption{Processing in the word layer: several processors operate in parallel. Their traversed paths are depicted ($\varnothing$: start nodes; $w_i$: word IDs; X: failure states).}
  \label{fig:rake}
\end{figure}
Once the character layer recognizes a word, e.g. \texttt{w6}, a new word node processor in the second layer is spawned (see upper left part of the figure; purple color).
If layer 1 then reports the next word \texttt{w7} (highlighted in green), processor 1 goes one step further in the graph now having a traversed path containing both words.
Additionally, another processor is spawned, starting directly with \texttt{w7}.
For this behavior, we use the metaphor of a rake: spawning another processor is like adding another tine to the rake.
Traversals in the word layer are only possible if the next detected word is a successor of the current one within any term of the FST, which, for example, is not the case when processor 2 tries to handle \texttt{w9}, or processor 4 tries to start directly with \texttt{w9}.
The latter means that there is no term in the FST starting with the word \texttt{w9}.
These two processors are then in a failure state (indicated by ``X'').
If there was a matching term in their traversed path, it is collected to be later processed by the voter.
If that is not the case, the failed processors may be removed from the rake.
The second case in which processors are removed, whether they are in a failure state or not, is after an explicit signal from the first layer, e.g. when reaching the end of a file or sentence.
Spawning additional processors to evaluate different possibilities in parallel especially originated from the latter.
Consider the case of interpreting a dot:
It could either indicate the end of a sentence (\textit{``Today, I met my Prof.''}), or an abbreviation (\textit{``Prof. Smith was also there.''}).
Thus, there is a forking in the second layer to evaluate both possibilities separately.
In theory, this could lead to endless forking, which is prevented by processors reaching failure states (i.e. given word sequences not matching any term) followed by their removal.
\\

\noindent\textbf{Real-Time Capability.}
Reading an input text of length $n$ characterwise yields a basic runtime complexity of $\mathcal{O}(n)$.
The same is true for processing $n$ characters in the first layer (at most $n$ transitions having a constant amount of operations; no backtracking needed).
The processing of a character may lead to the detection of a new word, which then triggers transitions in the word layer.
The number of these transitions depends on the number $p$ of processors (``tines in the rake'').
$p$ does not depend on $n$, but on the vocabulary, i.e. all words fed to the FST, especially $w_\text{max}$, the maximum number of words in all multi-word terms.
Although $p_{\text{max}}$ is constant for a given vocabulary, it may still be very large in worst case%
\footnote{
In worst case, a term consisting of $w_\text{max}$ words is read, whereas each subterm also exists in the vocabulary.
Moreover, for each of these subterms there is an additional variant ending with a dot.
This leads to forking after every word and a total amount of 
$p_{\text{max}} = \sum_{i=1}^{w_\text{max}}2^{i}$ processors before the first one of them fails and is removed.
}.
In practical scenarios however, $p \ll p_{\text{max}}$ can be assumed, since the vocabulary is only a tiny fraction of the power set of its words.
As a consequence, processors fail very fast due to given word combinations not matching any term in the FST.
Considering an additional constant amount of $c>0$ operations per processor in each transition of the second layer yields an upper limit of $c \cdot p_{\text{max}} \cdot n$.
Since $n$ is thus only multiplied with constants, the overall runtime complexity remains $\mathcal{O}(n)$.
Although the second layer's overhead is noticeable in practice (as we will see in Section \ref{sec:evaluation}), the overall runtime complexity is still linear and benefits our system's applicability in scenarios of real-time processing.
\\

\noindent\textbf{Inflection Tolerance.}
As mentioned before, to accept different lexical variations of terms, e.g. induced by inflection, we utilize information coming from connected ontologies as well as other language information sources.
Concerning the latter, we use a lemmatization table extracted from \textit{LanguageTool}%
\cref{ltmorph}, an open source proofreading software for several languages, which itself contains binary files of \textit{Morfologik}%
\footnote{\label{ltmorph}\url{https://github.com/languagetool-org} (uses \url{https://github.com/morfologik})} to look up part-of-speech data.
Such entries look as follows:
\begin{verbatim}
   künstlich     künstlich   ADJ:PRD:GRU
   künstliche    künstlich   ADJ:AKK:PLU:FEM:GRU:SOL
   künstlichem   künstlich   ADJ:DAT:SIN:MAS:GRU:SOL
\end{verbatim}
%
%
They contain the inflected form, its lemma as well as declension information like word class, case, number, gender, etc.
We additionally used \textit{Wiktionary}, a free wiki-based dictionary, whose data%
\footnote{\url{https://dumps.wikimedia.org/} (dump file of 2016-07-01)} we extracting using \textit{DKpro JWKTL}%
\footnote{\url{https://dkpro.github.io/dkpro-jwktl/}} \cite{ZeschMuellerGurevych2008}.
Nevertheless, there were still lots of words not covered by any of these sources, especially compound words like \textit{Forschungszentrum} (\textit{research center}).
To counter this, we additionally implemented heuristics like longest suffix matching to decompound words and use the inflected forms of the last part (if available).
In the case of \textit{Forschungszentrum}, these are the inflected forms of \textit{Zentrum} (\textit{center}), i.e. \textit{Zentrum}, \textit{Zentrums} and \textit{Zentren} as shown in Figure \ref{fig:fstht}.
Our tool is thus able to handle yet unknown words to a certain extent without user interaction.
In this regard, let us revisit the aforementioned 576 variations of the term DFKI.
As also mentioned, most of them are grammatically not correct.
Since we also want to handle yet unknown words, especially compound ones, while keeping the user interaction as low as possible (not asking for feedback), we decided to accept all variations obtained as the Cartesian product of all inflected forms of each of a term's words.
We assume that grammatically wrong variants do rarely occur in given texts and if they do, users will agree with the entity being recognized despite the misspelling.
Nevertheless, the question remains whether this decision considerably increases the false positive rate.
We will address this in Section \ref{sec:evaluation}.
To avoid actually harmful false positives of incorrectly inflected variants, we exploit additional ontological information like the type of an entity.
For example, the name of a person tolerates far less variants than the name of a topic.
Basically, we only allow a possessive/genitive case ``s'' at the end, like stated before.
As a consequence, our NE recognizer is actually not just a single MLFST, but a combination of several ones each having a different configuration.
Currently, there is one having higher and another one having lower tolerance.
The latter, for example, contains person names.
There is also an option to especially deal with acronyms:
They do not only require exact matches, moreover all characters need to be uppercase.
To further avoid non-meaningful variants, we only use adjective and noun information from the lemmatization table, which reduces ambiguities when not having thorough NLP information.
This is a compromise we can accept, since labels more often contain nouns and adjectives than verbs.

When processing input text, the different MLFST operate in parallel.
In the end, a voter receives, assesses, filters and finally returns their results.
Additionally, each MLFST has its own internal voter which assesses all results simultaneously present in a processing rake.
In the current implementation, these voters follow a strategy of only keeping the longest match.

%% file: content/main_evaluation.tex
\section{Evaluation}
\label{sec:evaluation}
%
\noindent\textbf{Setting.}
Besides finding out how fast our NE recognizer performs in practice, we were especially interested whether our design decisions (see Section \ref{sec:approach}) would lead to a considerable increase in false positives.
We were thus looking for large amounts of German natural language texts (prose) written by different people to test our approach.
The German Wikipedia meets this requirement but lacks ground truth data for the inflected forms present in these texts.
We therefore decided to only look at the wikilinks (see Figure \ref{fig:spotlight}, top section, blue words) and take them as a silver standard:
A human has annotated terms in the text (often in inflected form) with the label of their respective Wikipedia article (typically in basic form).
Figure \ref{fig:spotlight} also shows that users themselves decide which terms they annotate:
There are lots of entities (highlighted in green), which are not annotated although there is a Wikipedia article for them.
This is especially true for self-references, e.g. the term \textit{Aussagenlogik} is not annotated in ``its own'' article (i.e. the one about \textit{Aussagenlogik}).
Recognizers fed with such terms, would nevertheless find them, which has be considered when measuring precision.

Regardless of possible shortcuts, annotations are structured as follows:
the term appearing in the text and the name of the Wikipedia article it refers to (in the following also shortly called \textit{the link}) are written in double brackets separated by a pipe symbol, e.g. \texttt{[[Häuser|Haus]]} (plural form of \textit{house} appears in the text, whereas the article is labeled with the singular form).
Since inflection usually just changes one to four characters, the Levenshtein distance (LD) between term and link can help us identifying samples we could use to evaluate our approach.
Note that independent term-link-combinations like \texttt{[hometown|Eton]} or adjective-noun-combinations like \texttt{[entscheidbar|Entscheidbarkeit]}(\textit{decidable}) are undesirably also covered by such an LD-based heuristic.
On the other hand, this evaluation approach offers millions of inflection samples (we ran our tests on 3.9M articles having 50.4M annotations).

We downloaded German Wikipedia dump files%
\footnote{\url{https://dumps.wikimedia.org/} (dump file of 2016-11-01)} and used 3.9M article names as a basis for feeding our recognizers.
Disambiguation information in brackets like in \textit{``Berlin (Russland)''} (a village in Russia sharing its name with the German capital) were removed (this raises disambiguation issues as discussed later).
We also removed number-, symbol- and single-character-only labels, since they were not relevant for our investigations.
As ontological background information we used types%
\footnote{\url{https://downloads.dbpedia.org/3.9/de/instance_types_de.ttl.bz2}} coming from DBpedia, which were available for about 0.5M entities.
For types like person, city, film, etc., we applied a low tolerance strategy, whereas all other ones were treated with higher tolerance.
\\

\noindent\textbf{Evaluated NE Recognizers.}
We evaluated our MLFST approach against three baseline methods.
The first and most obvious one, \textit{StemFST}, was also implemented by us and uses the MLFST's character layer combined with the \textit{Lucene%
\footnote{\url{https://lucene.apache.org/}} German Stemmer}, which is based on \cite{Caumanns1999}.
The other methods are by Dlugolinsky et al. \cite{DlugolinskyNguyenLaclavik+2013}, who made two of their gazetteers available online%
\footnote{\url{http://ikt.ui.sav.sk/gazetteer/}}:
one based on hash-map multi-way trees (\textit{HMT}), and the other based on first child-next sibling binary trees (\textit{CST}).
Both produce the same results in terms of found NEs, but differ in memory consumption and runtime performance.

After filtering and editing as mentioned in the previous paragraph, we had slightly above 3.3M article names of the German Wikipedia that we fed to all four NE recognizers.
HMT and CST take these terms without further changes.
StemFST splits each term into words and reassembles it after stemming them.
Then it adds the altered term to its FST.
MLFST does the same but instead of stemming the words, it looks up (or tries to infer) their inflected forms.
Completely filled, the inner high-tolerance MLFST contained 8.5M character nodes and 3.5M word nodes, the low-tolerance part kept 1M and 0.4M nodes, respectively.
\\

\noindent\textbf{Results.}
All computations were performed on a notebook having an Intel Core i7-4910MQ 2.9 GHz CPU and 16 GB RAM, running on Windows 7 (64-bit).

\begin{figure}
\begin{minipage}[t]{0.45\textwidth}
  \centering
  \includegraphics[width=1\linewidth]{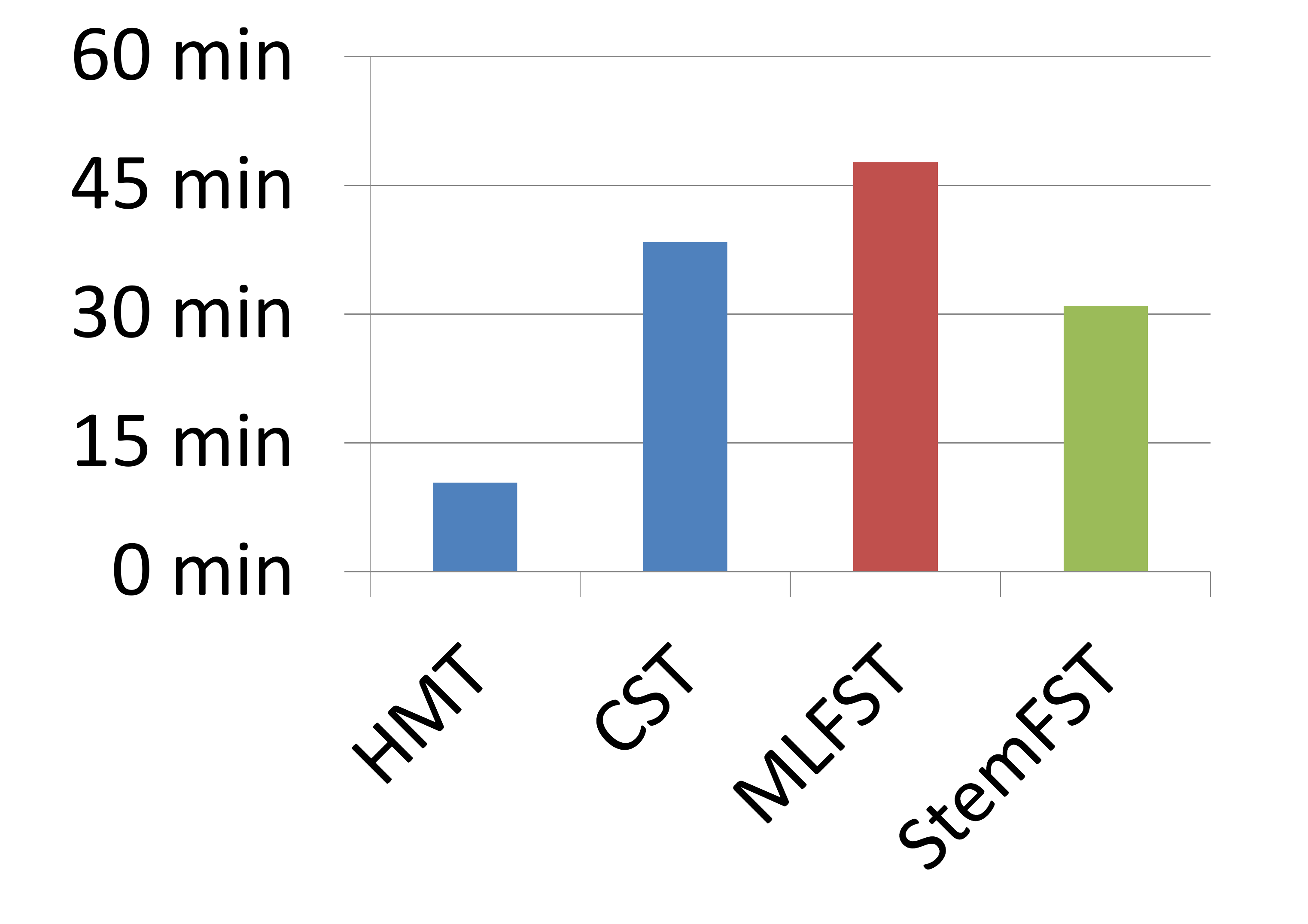}
  \vspace{-0.7cm}
  \captionof{figure}{Processing time}
  \label{fig:runtime}
\end{minipage}%
\hfill
\begin{minipage}[t]{0.45\textwidth}
  \centering
  \includegraphics[width=1\linewidth]{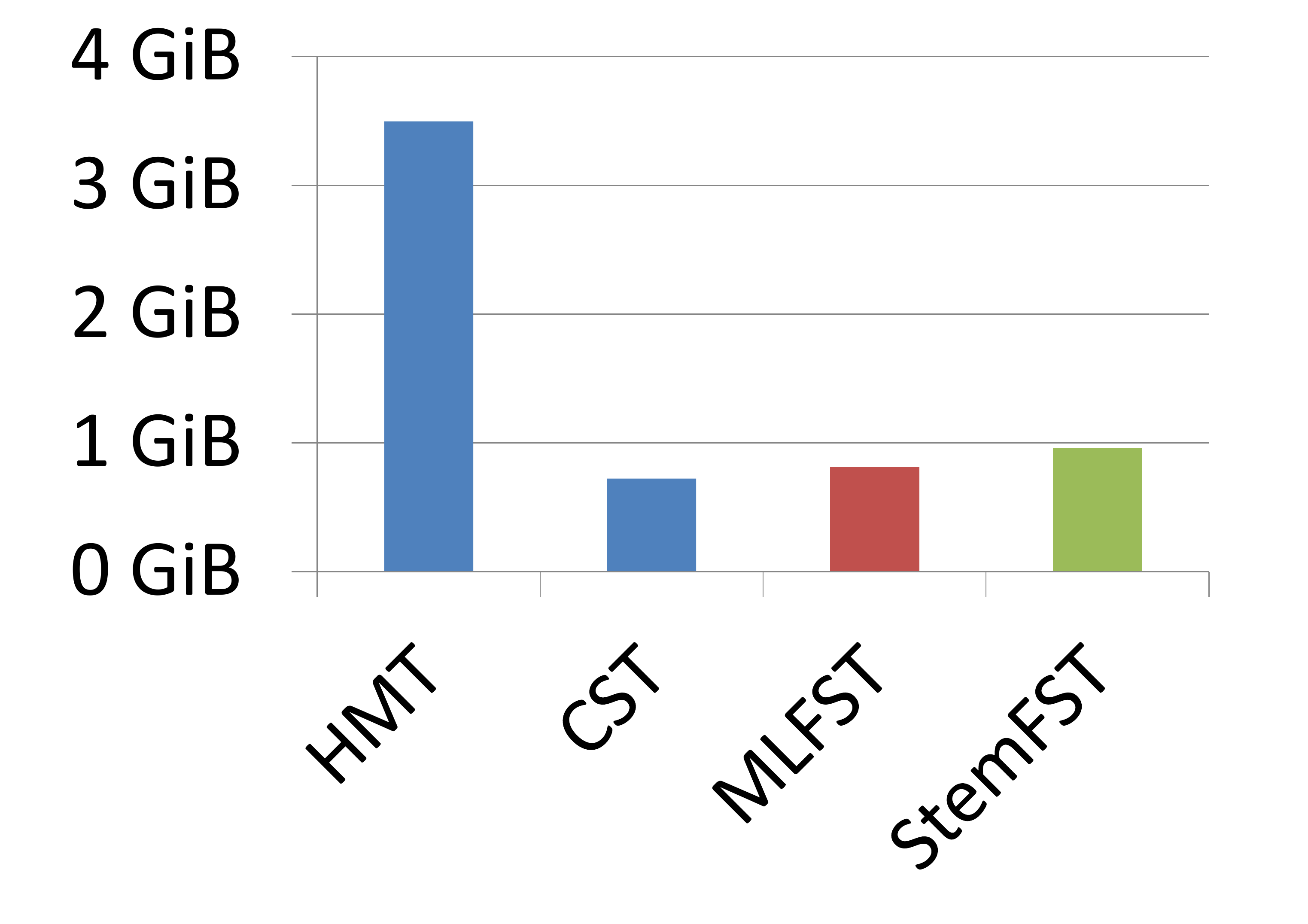}
  \vspace{-0.7cm}
  \captionof{figure}{Memory usage}
  \label{fig:memory}
\end{minipage}%
\end{figure}

HMT only needed 10.4 min for processing 3.9M articles (9.4B characters), while the others needed 31.0 to 47.7 min (see Figure \ref{fig:runtime}).
Figure \ref{fig:memory} shows that HMT trades memory efficiency for speed, since it is the only recognizer passing the 1 GiB mark by needing 3.5 GiB.
The others needed 0.72 to 0.96 GiB.

Let us next consider \textbf{recall}:
All recognizers reached values slightly below or above 70\%.
Figure \ref{fig:recall} additionally shows the results itemized by LD.
If term and link match exactly (\texttt{LD=0}, which is the case for 69\% of all annotations), all recognition rates are above 92\%%
\footnote{errors in the dump and imperfect parsing caused a slight decrease (100\% expected)}.
In LD ranges of \texttt{LD=1} to \texttt{LD=4} (11\% of all annotations), HMT/CST's recall is close to 0\%, whereas MLFST still has rates of 79\%, 66\%, 36\% and 8\%, respectively.
StemFST even has slightly higher rates.
Reaching recall near 100\% should not be expected, since not all variations are caused by inflection and their number decreases with increasing LD.
For higher LD values (\texttt{LD>4}, 21\% of all annotations), all recognition rates are close to 0\%.
\\

Concerning \textbf{precision}, we already mentioned the problem of how to measure it adequately.
We decided to calculate multiple values:
$P_{\text{O}}$ measures precision only for terms \textit{overlapping} with annotation positions, because only there we have ``ground truth'' data.
As shown in Figure 7, some found terms (purple highlighting) are not exactly matching the actual annotation (blue word, highlighted in green as the only true positive).
If terms are overlapping with the annotation, we count them as a false positive.
%
\begin{figure}
  \centering
  \includegraphics[width=1\columnwidth]{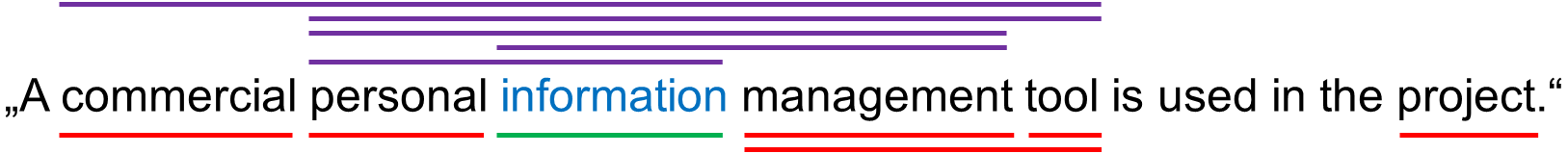}
  \caption{Example sentence to illustrate the different precision values}
  \label{fig:overlaps}
\end{figure}
%
$P_{\text{A}}$ counts \textit{all} terms not exactly matching as false positives, especially also the non-overlapping ones (red highlighting).
Since disambiguation was out of this paper's scope and there are labels belonging to more than 1000 instances (e.g. \textit{Jewish cemetery}), it makes a large difference whether or not we additionally count more than 1000 false positives for each true positive in a text.
We thus introduce $P_{\text{O}}^{*}$ and $P_{\text{A}}^{*}$, which count multiple entities having the same label only once.
%
\begin{figure}
\centering
\begin{minipage}[t]{0.45\textwidth}
  \centering
  \includegraphics[width=1\linewidth]{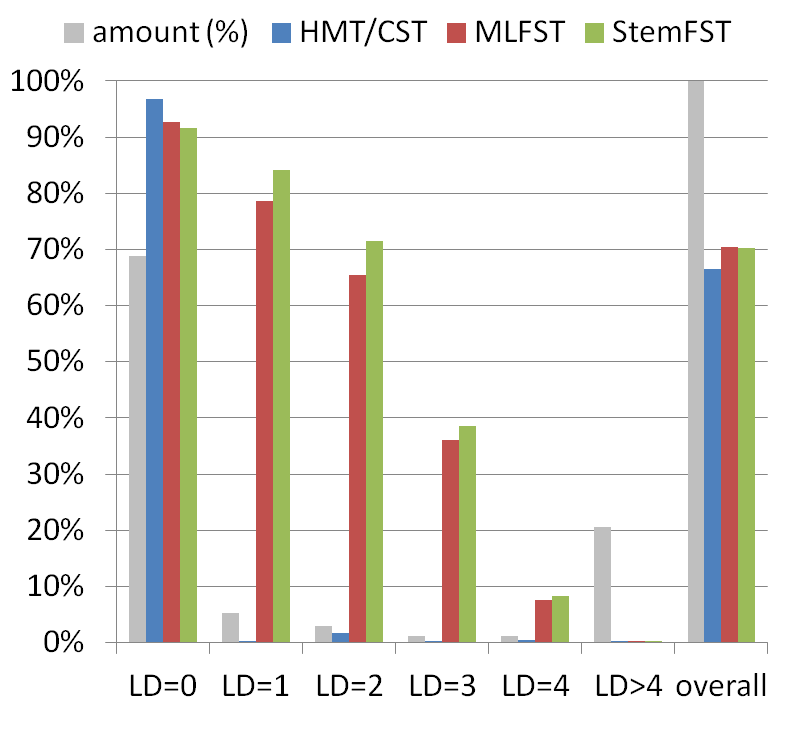}
  \captionof{figure}{Recall itemized by Levenshtein distance of term and link}
  \label{fig:recall}
\end{minipage}%
\hfill
\begin{minipage}[t]{0.45\textwidth}
  \centering
  \includegraphics[width=1\linewidth]{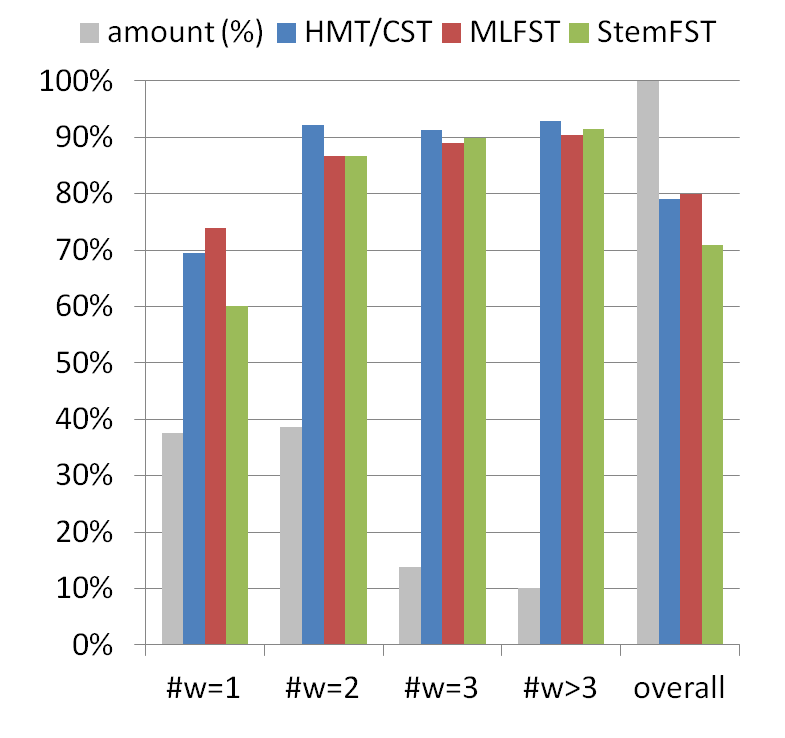}
  \captionof{figure}{Precision $P_{\text{O}}^{*}$ itemized by the terms' number of words (\#w)}
  \label{fig:precision1}
\end{minipage}
\end{figure}
%
$P_{\text{O}}^{*}$ is 79\% for HMT/CST and 80\% for MLFST, while StemFST only reaches 71\%.
Figure \ref{fig:precision1} additionally depicts $P_{\text{O}}^{*}$ itemized by the number of words a term consists of.
For multi-word terms, all approaches achieve values between 87\% and 92\%.
There is a remarkable difference for single word terms:
Here, stemming seems to be too rough causing terms to lose their specifity and StemFST to lose 14\% compared to MLFST, which performs best having 74\%.
The other overall precision values $P_{\text{O}}$, $P_{\text{A}}$ and $P_{\text{A}}^{*}$ are shown in Figure \ref{fig:precision2}.
They are far lower than $P_{\text{O}}^{*}$ due to the aforementioned reasons.
However, in a short experiment, in which students annotated some randomly chosen articles manually, we observed values for $P_{\text{A}}^{*}$ that were similar to $P_{\text{O}}^{*}$ above.
We thus have a slight indication that $P_{\text{A}}^{*}$ above heavily underestimates our algorithm's precision.
Finally answering one of our initial research questions: 
the false positive rate of MLFST is not considerably higher (in some cases even lower) than with the other recognizers.
\\

Regarding \textbf{runtime performance}, MLFST and StemFST process between 3281 and 5048 characters per millisecond and are thus comparable to CST as illustrated in Figure \ref{fig:chpms}.
%
\begin{figure}
\centering
\begin{minipage}[t]{0.45\textwidth}
  \centering
  \includegraphics[height=5.7cm]{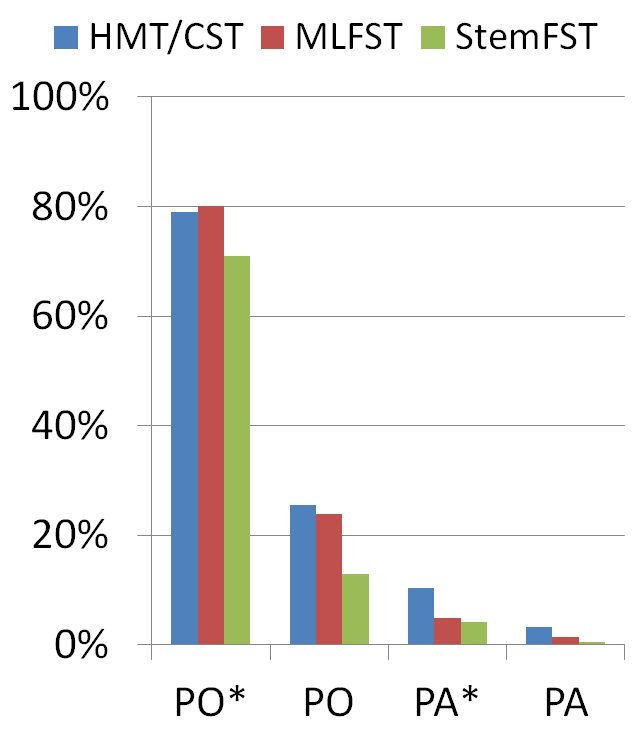}
  \captionof{figure}{Precision: $P_{\text{O}}^{*}$, $P_{\text{O}}$, $P_{\text{A}}^{*}$,  $P_{\text{A}}$}
  \label{fig:precision2}
\end{minipage}%
\hfill
\begin{minipage}[t]{0.45\textwidth}
  \centering
  \includegraphics[height=5.7cm]{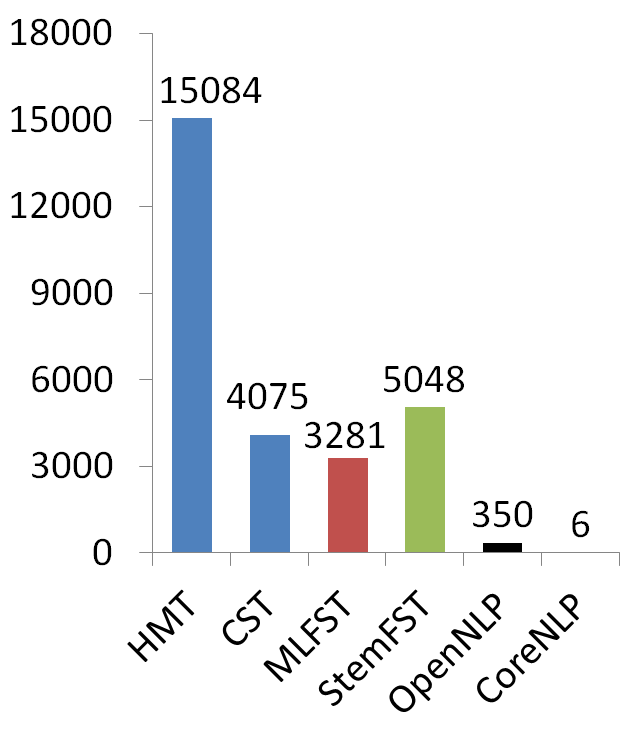}
  \captionof{figure}{Processed characters per ms}
  \label{fig:chpms}
\end{minipage}
\end{figure}
%
HMT is about three times faster at the expense of memory consumption (see Figure \ref{fig:memory}).
All tested recognizers are by orders of magnitude faster than basic NLP pipelines.
We tested OpenNLP%
\footnote{\url{https://opennlp.apache.org/}} and CoreNLP using a basic pipeline consisting only of a tokenizer, sentence splitter and part-of-speech tagger.
Although no NER-specific analyzers like noun chunkers or type classifiers were added yet, their processing time was already out of our targeted range.
Running the basic pipeline on all 3.9M articles would presumably have taken about 18 days in the case of CoreNLP, for example.

%% file: content/main_conclusion.tex
\section{Conclusion \& Outlook}
\label{sec:conclusion}
\noindent
In this paper, we presented an ontology-based NER approach that is comparably fast as available high speed methods while outperforming them in the recognition of terms that lexically vary slightly, e.g. induced by inflection.
We were thus able to narrow the quality gap to more sophisticated but also much slower NLP pipelines a bit more without loosing to much runtime performance.

In the future, we plan to additionally incorporate StemFST into MLFST, since its recall was slightly better for multi-word terms.
Additionally, we could add more layers scanning for patterns like phrases that indicate todos or appointments, Hearst patterns \cite{Hearst1992}, etc.
There is also much potential for improving the language capabilities of our approach, e.g. improved rules and heuristics or multi-language support.
Last not least, we plan to incorporate disambiguation mechanisms by exploiting the explicated user context available in our system.
\\

%% file: content/frontback_acks.tex
\noindent
\textbf{Acknowledgements.}
This work was funded by the Deutsche Forschungsgemeinschaft (DFG, German Research Foundation) -- DE 420/19-1.

\noindent
We would also like to thank Sven Hertling, Jörn Hees, Erfan Shamabadi, Oleksii Kotvytskyi and Tim Sprengart for their contributions in this project's early and late phase, respectively.